\documentclass[10pt,twocolumn,letterpaper]{article}
\usepackage{wacv}
\usepackage{times}
\usepackage{epsfig}
\usepackage{graphicx}
\usepackage{amsmath}
\usepackage{amssymb}

\wacvfinalcopy 

\def\wacvPaperID{410} 

\ifwacvfinal\fi
\begin{document}

\title{\textbf{C\textsuperscript{2}MSNet}: A Novel approach for single image haze removal}

\author{Akshay Dudhane\textsuperscript{1} \hspace{2cm} Subrahmanyam Murala\textsuperscript{2}\\
Computer Vision and Pattern Recognition Lab\\
Indian Institute of Technology Ropar\\
India\\
{\tt\small \textsuperscript{1}2017eez0001@iitrpr.ac.in}\\
{\tt\small \textsuperscript{2}subbumurala@iitrpr.ac.in}
}

\maketitle
\ifwacvfinal\thispagestyle{empty}\fi

\begin{abstract}
   Degradation of image quality due to the presence of haze is a very common phenomenon. Existing DehazeNet \cite{cai2016dehazenet}, MSCNN \cite{ren2016single} tackled the drawbacks of hand crafted haze relevant features. However, these methods have the problem of color distortion in gloomy (poor illumination) environment. In this paper, a cardinal (red, green and blue) color fusion network for single image haze removal is proposed. In first stage, network fusses color information present in hazy images and generates multi-channel depth maps. The second stage estimates the scene transmission map from generated dark channels using multi channel multi scale convolutional neural network (McMs-CNN) to recover the original scene. To train the proposed network, we have used two standard datasets namely: ImageNet \cite{deng2009imagenet} and D-HAZY \cite{ancuti2016d}. Performance evaluation of the proposed approach has been carried out using structural similarity index (SSIM), mean square error (MSE) and peak signal to noise ratio (PSNR). Performance analysis shows that the proposed approach outperforms the existing  state-of-the-art methods for single image dehazing.
\end{abstract}

\section{Introduction}

Low visibility in bad weather is a foremost problem in the computer vision. Appearance of the object in a human vision system requires the reflection of the atmospheric light from the object surface to the human eyes. However, presence of atmospheric particles (haze, fog etc.) in the transmission medium causes visual degradation in the quality of image. Most of the computerized systems assume that the input images are captured in clear atmosphere. Whereas, outdoor images undergo visibility degradation, especially in the morning (in presence of haze, fog etc.). So, to remove atmospheric effect computer vision algorithm related to outdoor imaging should be acquainted about the outdoor weather effect (haze). To achieve haze removal there is a dire need to develop an algorithm which could enhance the image quality in real time.

\section{Related Work}
Various approaches had been proposed such as 3D geometrical model \cite{609419}, polarisation filters \cite{schechner2001instant, 1640996}, fusion of multiple images of same scenery \cite{narasimhan2002vision, nayar1999vision} etc. for a single image haze removal. However, in real time applications, collecting multiple images of same scenery is not always feasible. Further, with a simple and convincing assumptions, dehazing models were proposed by \cite{ancuti2010fast, fattal2008single, he2011single, huang2014visibility, tan2008visibility, zhu2014single}.
Tan \etal \cite{tan2008visibility} assumed contrast of the haze free image must be higher than the hazy scene. So, with this aforementioned assumption they removed haze by maximizing the local contrast of the hazy image. This method \cite{tan2008visibility} fails and create blocking artifacts when there is a depth discontinuity in the hazy image.
\par Another effective local approach, dark channel prior (DChP) was proposed by He \etal \cite{he2011single}. According to DChP, most of the local image patches in outdoor haze free colour images are comprised of some pixels (dark pixels) having very low intensity among at least one of their colour channels. According to Codruta \etal \cite{ancuti2010fast} haze density can be predicted not only with the cardinal channel (RGB) but also in the HSV color space. They used the hue disparity between hazy image and its semi inverse image to detect the haze content. Huang \etal \cite{huang2014visibility} proposed visibility restoration (VR) algorithm to reduce color distortion and deal with complex structures which incorporates the edge information in the transmission map obtained using DChP. Alternatively, Zhu \etal proposed a colour attenuation prior (CAP) \cite{zhu2014single} which estimates the depth map through HSV colour space.
\par Approaches discussed above witnessed the correlation of haze content with the color models and their advantages. Hence, learning the combination of all the assumption would anticipate the haze density more accurately. Because, single assumption does not guaranteed to be precise always, which may fail in extreme vague environment and complex structures. So, to develop a more robust system, \cite{cai2016dehazenet, ren2016single} adopted convolution neural networks to learn the mapping between input hazy image and corresponding transmission map. However, \cite{cai2016dehazenet, ren2016single} introduces the color distortion in gloomy (poor illumination) environment. Because, these networks loose the cardinal channel information in the first convolutional layer itself. Also, assumption made by the He \etal \cite{he2011single} allows to split the cardinal channels and anticipate the haze spread. 
\par Inspired from \cite{cai2016dehazenet, he2011single, ren2016single} in this paper, we have proposed a new network to integrate the color information with respect to the gloominess in the environment and multi-scale filter bank to learn the pixel to pixel mapping between input hazy image and transmission map. We have trained this network on 3,50,000 variety of patches. Also, proposed network is tested on more than one thousand real as well as synthetic images. 

\section{Background}
In this section we have highlighted the approach proposed by He \etal \cite{he2011single}: 
\subsection{Optical Model}
Optical model (OM) is based on the physical properties of the transmission of light through the air medium. Many computer vision algorithms make use of the optical model to describe the image formation. Especially, in haze removal many researchers used OM to reconstruct the original scene \cite{he2011single, tan2008visibility}. As given in \cite{he2011single}, Eq. 1 describes the optical model.
\begin{equation}
\label{eq}
I\left( x \right)=R\left( x \right)Tr\left( x \right)+A\left( 1-Tr\left( x \right) \right)
\end{equation}
where, \textit{I(x)} is the pixel intensity of hazy image, \textit{R(x)} is the scene radiance, \textit{Tr(x)} is the transmission medium and \textit{A} is the global atmospheric light.\\
Procedure to estimate the transmission map($Tr$) and global airlight($A$) using dark chennel prior \cite{he2011single} is given below.  
\subsection{Dark channel prior}
Though there are many existing drawbacks, assumption made by the He \etal \cite{he2011single} exhibits a strong solution to estimate the depth information and haze density. Keeping dark channel prior as a basis one can learn a strong haze features to restore the original scene. As discussed in section 2, with the key assumption He \etal \cite{he2011single} estimated the scene depth using Eq. 2,
\begin{equation}
\label{eq}
D\left( x \right)=\underset{l\in P\left( x \right)}{\mathop{\Gamma }}\,\left( \underset{c\in \left\{ r,g,b \right\}}{\mathop{\Gamma }}\,{{I}^{c}}\left( l \right) \right)
\end{equation}
where, \textit{D(x)} represents the dark channel, $\Gamma$($\cdot$) represents the minimum filter, \textit{P(x)} belongs to the local patch cantered at location \textit{x} having size of $15\times 15$, \textit{c} is the colour channel of the hazy image, \textit{I} is the hazy image.
\par It can be understood from the Eq. 2, if \textit{I} is the haze free image then \textit{D(x)} will be closer to zero. Basically, dark channel grasps the color information and anticipates the haze density. So, assumption made by the He \etal allows us to split the cardinal channels and fuse them to predict/learn the haze spread/density. 
\subsection{Transmission map}
Transmission map estimated using Eq. 3,
\begin{equation}
\label{eq}
Tr\left( x \right)=1-\eta \underset{c}{\mathop{\Gamma }}\,\left( \underset{l\in P\left( x \right)}{\mathop{\Gamma }}\,\left( \frac{{{I}^{c}}\left( P \right)}{{{A}^{c}}} \right) \right)
\end{equation}
where, \textit{Tr(x)} is a estimated transmission map, \textit{$\eta$} is a haze constant is set to 0.95, $\Gamma$($\cdot$) represents a minimum filter, \textit{P(x)} belongs to the local patch cantered at location \textit{x} and patch size is taken as $15\times 15$, \textit{c} is the colour channel of the hazy image and \textit{I} is the hazy image and \textit{A} is an estimated global air light.\par He \etal \cite{he2011single} used dark channel to predict the airlight. Locations of $0.1\%$ brightest pixel from the dark channel are accessed from hazy image and mean of these intensities taken as a global air light. Now, scene radiance (haze free image) \textit{R(x)} can be easily recovered by using Eq. (1).

\section{Proposed Method}
Even though the DChP predicts the haze density, it fails in presence of dense haze and introduces the color distortion. Because, single depth map can not estimate the spread of the haze accurately. Optical model suggests that the precise estimation of depth/transmission map is the key to recover the scene radiance. Hence, to recover the scene radiance, we have designed a novel $cardinal ~color ~fusion ~multi-scale ~CNN ~(C\textsuperscript{2}MSNet)$. Proposed network is divided into two stages, in first stage, we estimate the multiple depth channels (MDCh)\footnote {C\textsuperscript{2}MSNet is designed to learn the different combinations of possible priors, it generates multiple depth maps. Here multiple depth map word indicates the haze relevant features.} using cardinal color fusion CNN ($C^2$-$NN$) and in second stage, multi-channel multi-scale CNN is used to learn the haze density/spread with respect to the transmission map.
\subsection{Cardinal color fusion CNN (C\textsuperscript{2}-NN)}
As discussed in section 2, haze is proportional to the intensity variation in various color channels. This observation lead us to fuse the color information and to design a novel network for estimation of depth maps. Proposed \textit{$C^2$-$NN$} consists of convolution and pooling layers. Figure 1(a) shows the network architecture. Proposed \textit{$C^2$-$NN$} divided into three sub-blocks namely: channel-wise -$\{convolution, concatenation ~and ~pooling\}$. Assumptions made by the \cite{ancuti2010fast, fattal2008single, he2011single, huang2014visibility, tan2008visibility, zhu2014single} allow us to learn the haze spread by \textit{spilt and fuse operation} on the color channels. In order to fuse the color information, initially we split the cardinal channels and applied parallel convolution filter banks of size \textit{$3\times 3\times 1\times 32$}. Further, channel-wise concatenation (CwC) is performed on filter response using Eq. (4).
\begin{figure*}[t]
\begin{center}
   \includegraphics[width=0.75\linewidth]{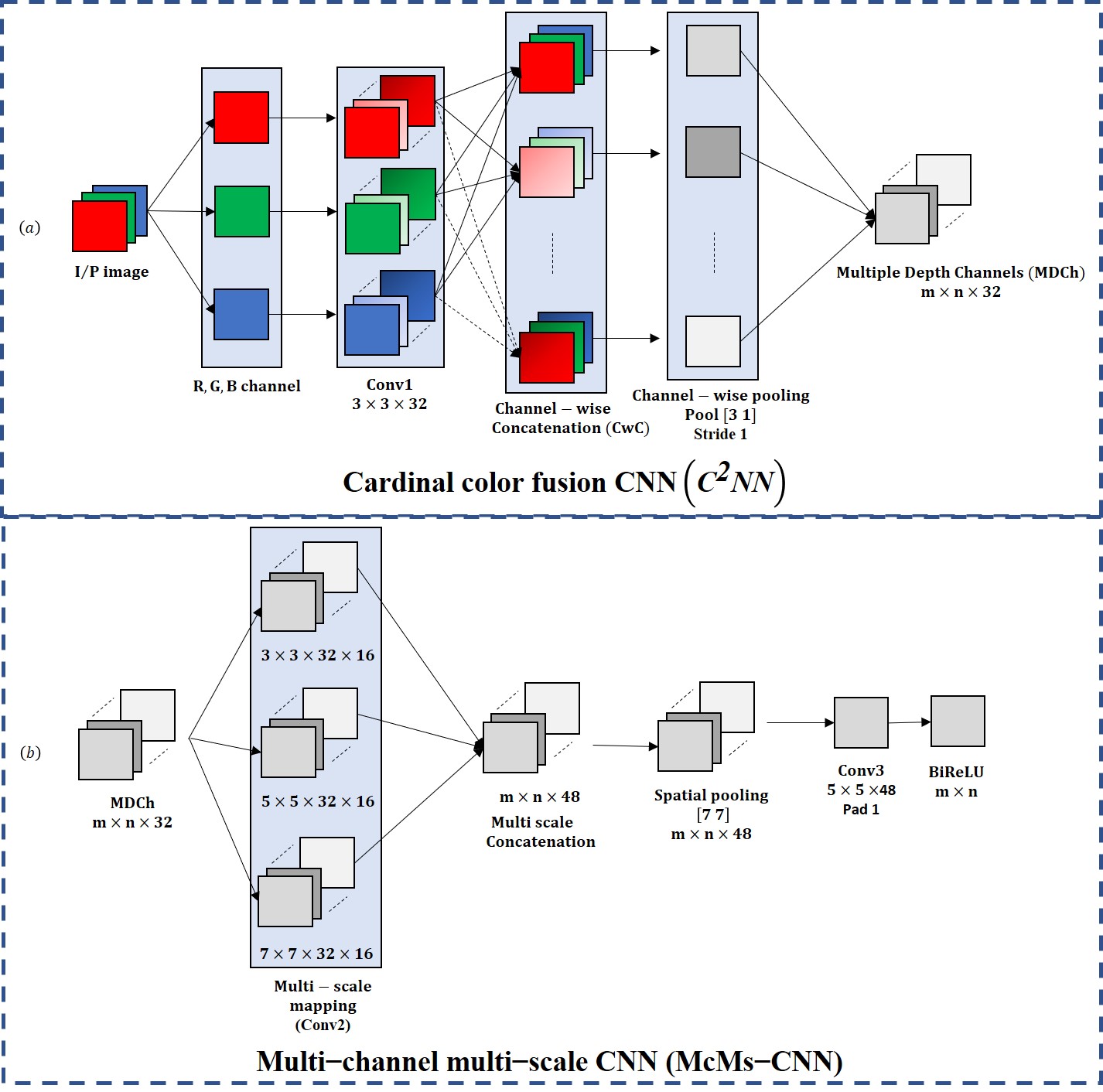}
\end{center}
   \caption{Proposed C\textsuperscript{2}MSNet: (a) novel cardinal color fusion CNN (b) multi-channel multi-scale CNN.}
\label{fig:1}
\end{figure*}

\begin{equation}
\label{eq}
Cw{{C}_{i}}=\Cap \left\{ \Psi _{i}^{R},\,\Psi _{i}^{G},\,\Psi _{i}^{B} \right\}
\end{equation}
where, \textit{$\Cap\{\cdot\}$} represents the concatenation operation, \textit{$\Psi_{i}^R$}, \textit{$\Psi_{i}^G$} and \textit{$\Psi_{i}^B$} represent the convolution responses of cardinal channels respectively and \textit{i $\in$ $\{1,2,..,32\}$} are the number of convolution filters.
\par Finally, the channel-wise max pooling is performed on output of the concatenation layer. Output dimensions of the \textit{$C^2$-$NN$} are \textit{$m\times n\times 32$}. Whereas, $m$ and $n$ represent the image height and width respectively.

\subsection{Multi-channel multi-scale CNN (McMs-CNN)}
He \etal assumed 15$\times$15 sized patch to obtain a dark channel prior. However, this fixed size patch makes DChP scale variant and causes halo effects at depth discontinuity. So, to make proposed network scale invariant, we designed multi-scale filter banks followed by spatial pooling and non-linear activation unit to retain the structural information. The output of \textit{$C^2$-$NN$} is given to the McMs-CNN which consists of five layers as follows: $\{multi scale ~convolution\rightarrow multi scale ~concatination\\ \rightarrow~pooling\rightarrow ~convolution\rightarrow ~BiRelu\}$.
\par To collect the multi-scale information, we have used three different filter banks, each having $16$ filters: \textit{$3\times 3\times 16;~ 5\times 5\times 16 ~and ~~7\times 7\times 16$}. The multi scale convolution layer refines the MDCh and the multiscale-concatenation operation is performed to conjoin the haze features at coarse level which results in $48$ feature maps. Further, the combination of spatial pooling and convolution layer have been used to extract the local haze relevant features. The spatial max pooling is performed within $7\times 7$ region having \textit{stride} of one. The convolution ($conv3$) layer consists of $48$ filters of size $5\times 5$ resulting $5\times 5\times 48$ filter bank. Finally, the non-linear activation function (BiReLU) \cite{cai2016dehazenet} has been used to anticipate the transmission channel.

\subsection{Training of C\textsuperscript{2}MSNet}
To construct the training set of hazy images, we have used D-HAZY \cite{ancuti2016d} dataset which is generated from NYU depth indoor image collection \cite{silberman2012indoor}. D-HAZY \cite{ancuti2016d} contains depth map for each indoor hazy image. We have generated $2,00,000$ hazy patches (size $64\times 64$) along with respective depth maps from D-HAZY \cite{ancuti2016d} dataset. Even though this database consists huge number of hazy images with depth maps, it does not consists outdoor natural scenes. So, to learn the haze spread in natural hazy environment we have extracted $1,50,000$ patches (size $64\times 64$) from ImageNet \cite{deng2009imagenet} dataset. The hazy image model given by Eq. 1 is used to generate the synthetic hazy patches from ImageNet database. In total, we have generated 3,50,000 hazy patches with respective depth and transmission maps. Further, these hazy patches are randomly sampled and $1,50,000$ patches were taken out to train the proposed C\textsuperscript{2}MSNet and $2,00,000$ patches for validation of the network.
\par The C\textsuperscript{2}MSNet is initialized with random weight parameters and trained using stochastic gradient descent (SDG) back propagation algorithm with a batch size of 64 and learning rate of 0.002. The mean square error (MSE) is used as the loss function. The weight parameters of C\textsuperscript{2}MSNet are updated in 18 epochs on PC with 4.20 GHz Intel Core i7 processor and NVIDIA GTX 1080 8GB GPU. Figure 2 shows the training process of C\textsuperscript{2}MSNet with aforementioned settings. Also, we have calculated the time complexity of different methods for single image haze removal. We run the different algorithms on the same machine as described above. Table 1 shows the execution time taken by various methods for single image haze removal. From Table 1, it is clear that the proposed network is computationally efficient as compared to existing algorithms\footnote {We have used executable codes of different methods made available by the respective authors on their websites. The image size considered for the evaluation is 100$\times$100$\times$3.}. 

\begin{figure}[t]
\begin{center}
   \includegraphics[width=1\linewidth]{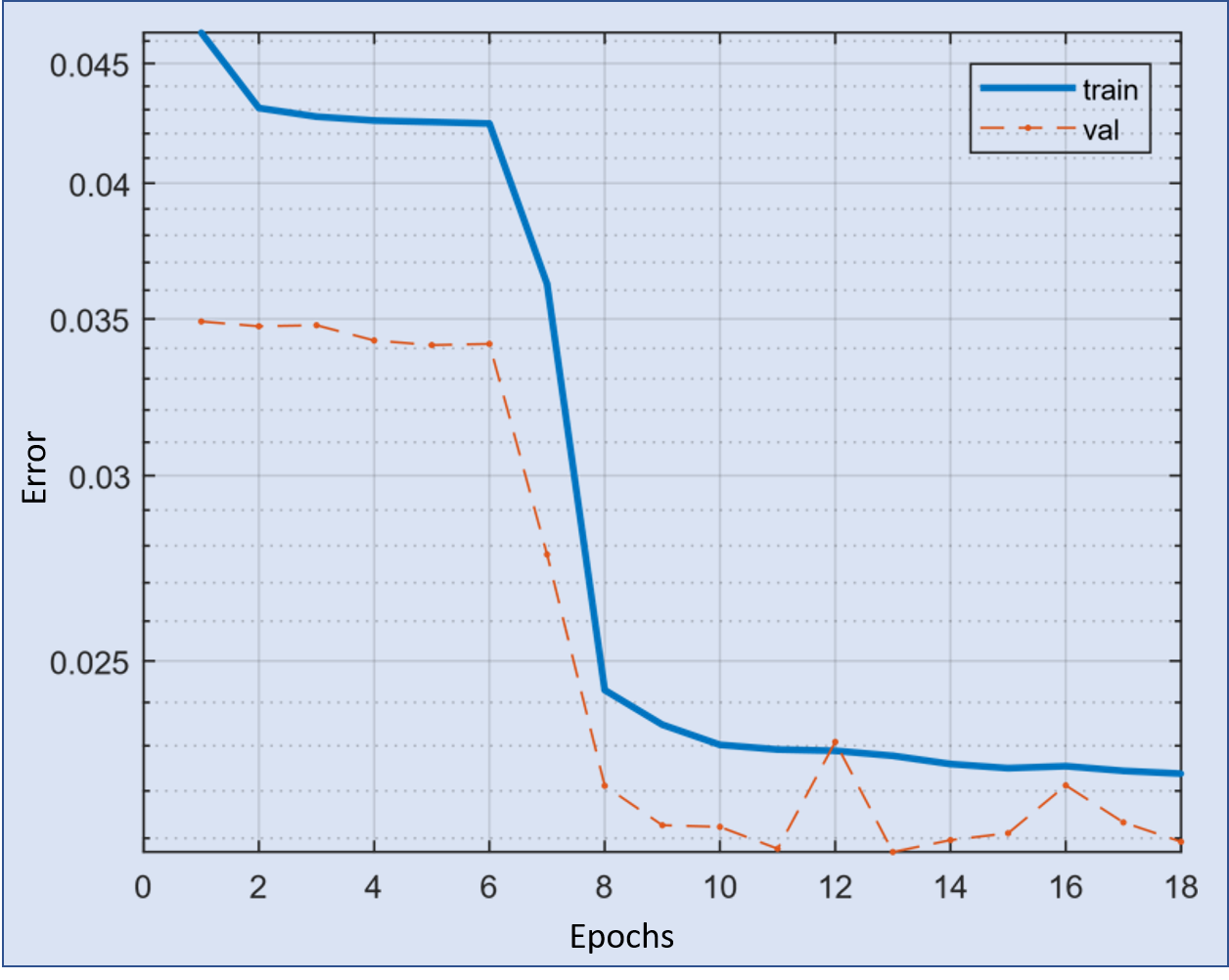}
\end{center}
   \caption{C\textsuperscript{2}MSNet training process with mean square error (MSE) as an error function.}
\label{fig:2}
\end{figure}

\section{Experimental results and discussion}
In this section, the proposed de-hazing network is tested with both qualitative and quantitative analysis. The structural similarity index (SSIM), mean square error (MSE) and peak signal to noise ratio (PSNR) are considered for the quantitative analysis. We have categorized the experiments into three parts: indoor synthetic hazy images with depth maps, synthetic outdoor images and real hazy images without depth maps.
\begin{figure*}[t]
\begin{center}
\includegraphics[width=0.9\linewidth]{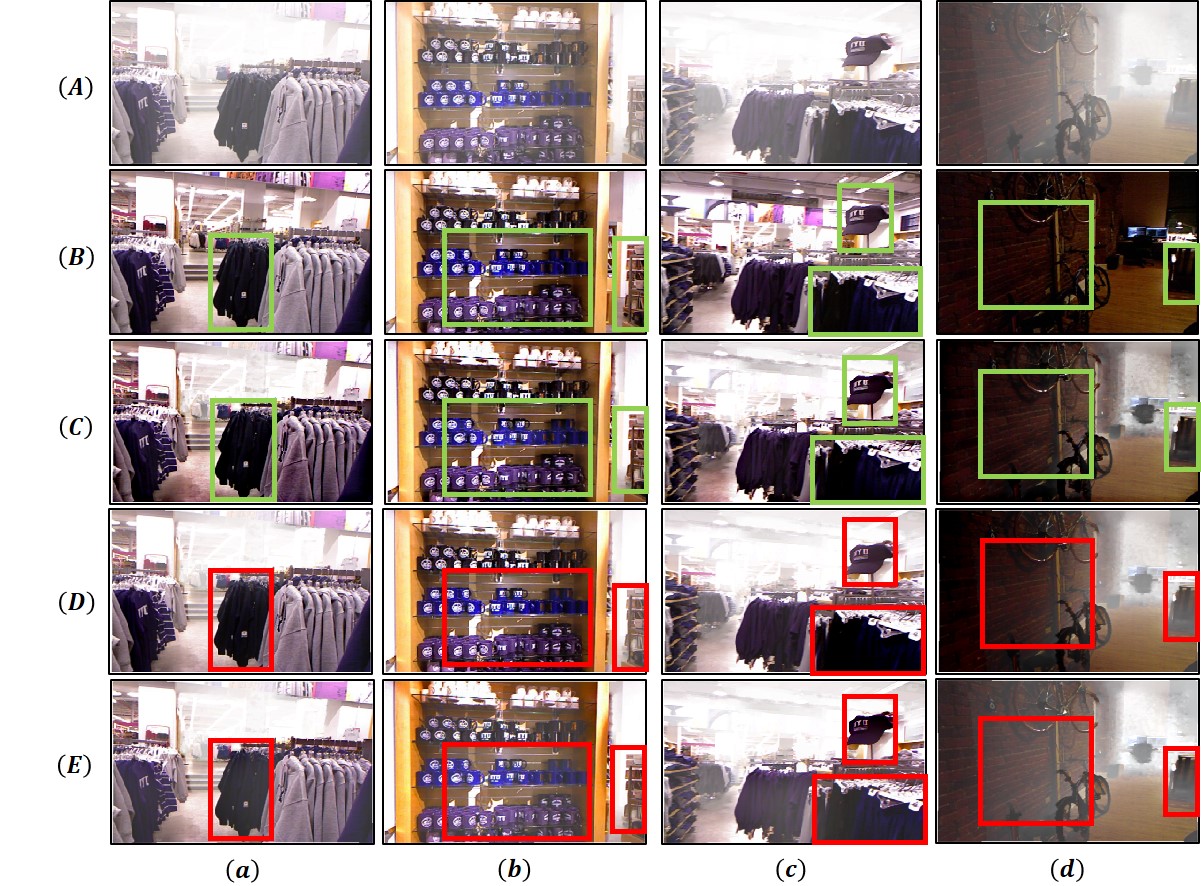}
\end{center}
   \caption{Result comparison between state-of-the-art methods and proposed C\textsuperscript{2}MSNet for single image haze removal: (A) Sample hazy images from D-HAZY \cite{ancuti2016d} database (B) Ground truth (C) Proposed method (D) DehazeNet\cite{cai2016dehazenet} (E) MSCNN \cite{ren2016single}. Please magnify figure to see color distortion in detail.}
\label{fig:3}
\end{figure*}

\subsection{Exp. 1: Indoor synthetic hazy images}
As mentioned in section 4.3, D-HAZY is a benchmark dataset for single image haze removal. It contains 1,449 hazy images along with their depth maps respectively.
\subsubsection{Quantitative analysis}
\par The performance evaluation of the proposed network has been addressed by testing 642 randomly sampled images from D-HAZY \cite{ancuti2016d} dataset. Also, we have compared the performance of C\textsuperscript{2}MSNet with existing state-of-the-art methods \cite{cai2016dehazenet, he2011single, ren2016single, zhu2015fast} as shown in Table 2. The considered evaluation measures are average values of SSIM, MSE and PSNR over 642 images. From Table 2, it is evident that the C\textsuperscript{2}MSNet outprforms the other existing methods for single image haze removal on D-HAZY \cite{ancuti2016d} dataset.

\begin{table}[b]
\begin{center}
\caption{average time taken by various methods to process a single image.}
\vspace{2mm}
\begin{tabular}{|l|c| c|}
\hline
Method & Time (sec.) & Platform\\
\hline\hline
He \etal \cite{he2011single} & 0.17 & MATLAB\\
Zhu \etal \cite{zhu2015fast} & 0.20 & MATLAB\\
Cai \etal \cite{cai2016dehazenet} & 0.09 & MATLAB\\
Ren \etal \cite{ren2016single} & 0.14 & MATLAB\\
\textbf{Proposed method}  & \textbf{0.072} & MATLAB\\
\hline
\end{tabular}
\end{center}

\end{table}

\begin{table}[b]
\begin{center}
\caption{Quantitative analysis of single image haze removal on D-HAZY \cite{ancuti2016d} dataset.}
\vspace{2mm}
\begin{tabular}{|l|c c c|}
\hline
Method & SSIM & MSE & PSNR \\
\hline\hline
He \etal \cite{he2011single} & 0.8280 & 0.0406 & 14.4794 \\
Zhu \etal \cite{zhu2015fast} & 0.8145 & 0.0379 & 14.7497 \\
Cai \etal \cite{cai2016dehazenet} & 0.8199 & 0.0391 & 14.7265\\
Ren \etal \cite{ren2016single} & 0.7865 & 0.0484 & 13.6193 \\
\textbf{Proposed method}  & \textbf{0.8480} & \textbf{0.0239} & \textbf{16.5808}\\
\hline
\end{tabular}
\end{center}

\end{table}

\begin{figure*}[t]
\begin{center}
\includegraphics[width=0.9\linewidth]{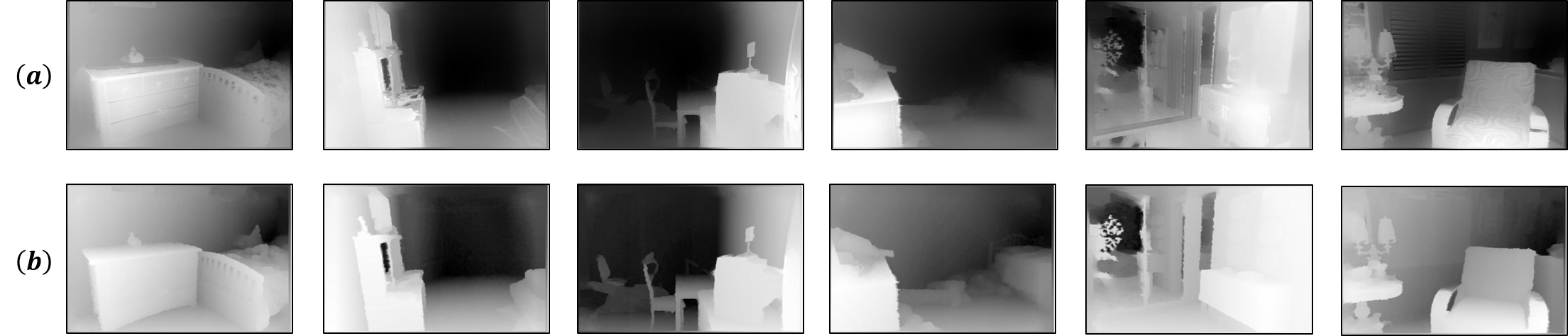}
\end{center}
   \caption{Visual comparison between ground truth and estimated transmission maps using proposed C\textsuperscript{2}MSNet on D-Hazy \cite{ancuti2016d} database. (a) Ground truth transmission map (b) Estimated transmission map using proposed method.}
\label{fig:4}
\end{figure*}

\begin{figure*}[t]
\begin{center}
\includegraphics[width=0.9\linewidth]{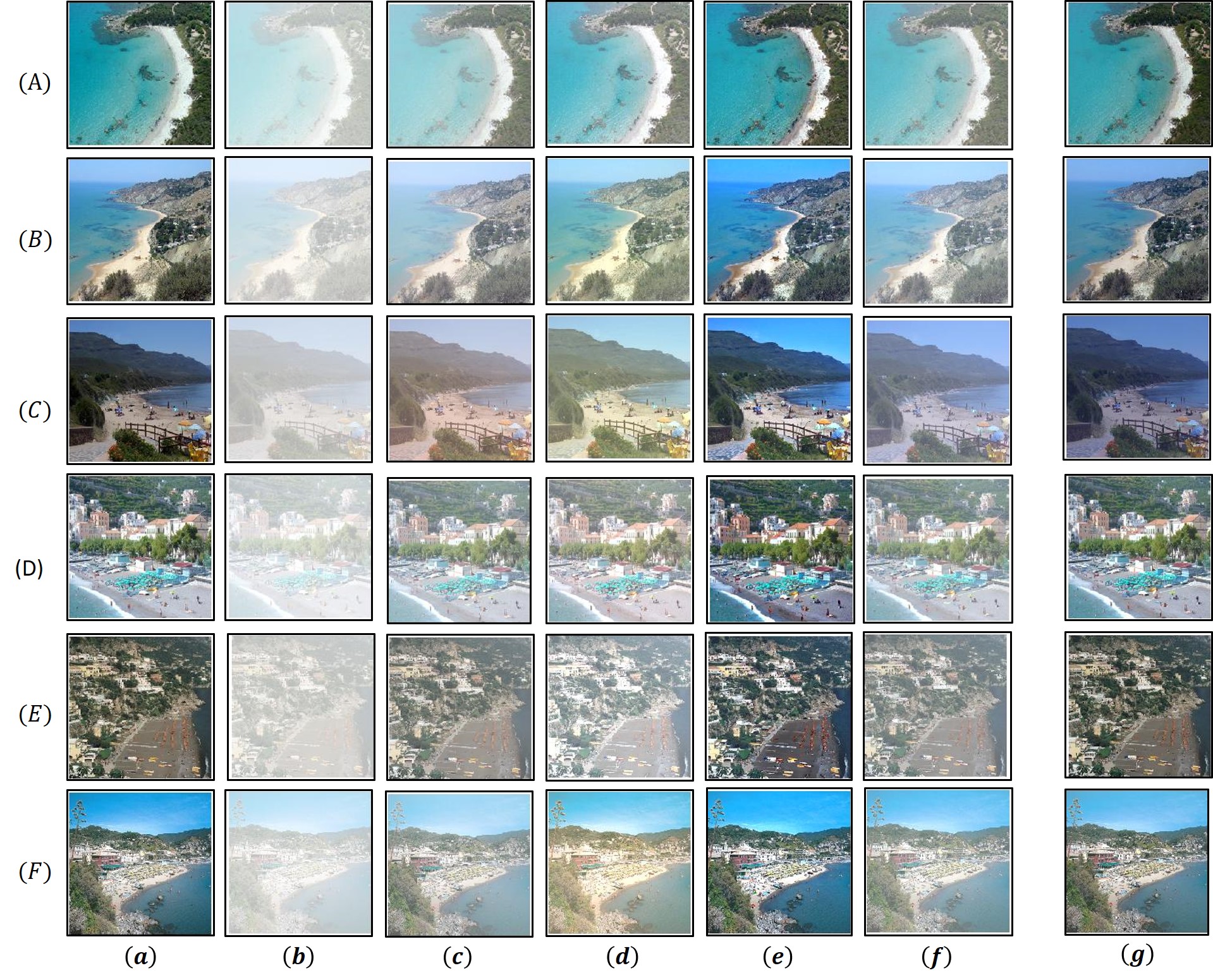}
\end{center}
   \caption{Visual comparison for outdoor synthetic image haze removal on OHI dataset.(a) Haze free sample images (b) synthetic hazy image (c) DehazNet\cite{cai2016dehazenet} (d) Ren \etal \cite{ren2016single} (e) He \etal \cite{he2011single} (f) Zhu \etal \cite{zhu2015fast} (g) Proposed method.}
\label{fig:5}
\end{figure*}

\subsubsection{Qualitative analysis}
\par Figure 3 illustrates the synthetic hazy images and their corresponding scene radiance, recovered using proposed network and existing \cite{cai2016dehazenet, ren2016single} methods. Figure 3(D-E) witnessed to the color distortion in recovered scene radiance with the DehazeNet \cite{cai2016dehazenet} and MSCNN \cite{ren2016single} approaches. Specfically, in Figure 3(c,D-E), as input image contains little gloomy illumination in marked region, \cite{cai2016dehazenet, ren2016single} introduce color distortion in such a manner that edges in that region are not separable to
\begin{figure*}[t]
\begin{center}
\includegraphics[width=0.98\linewidth]{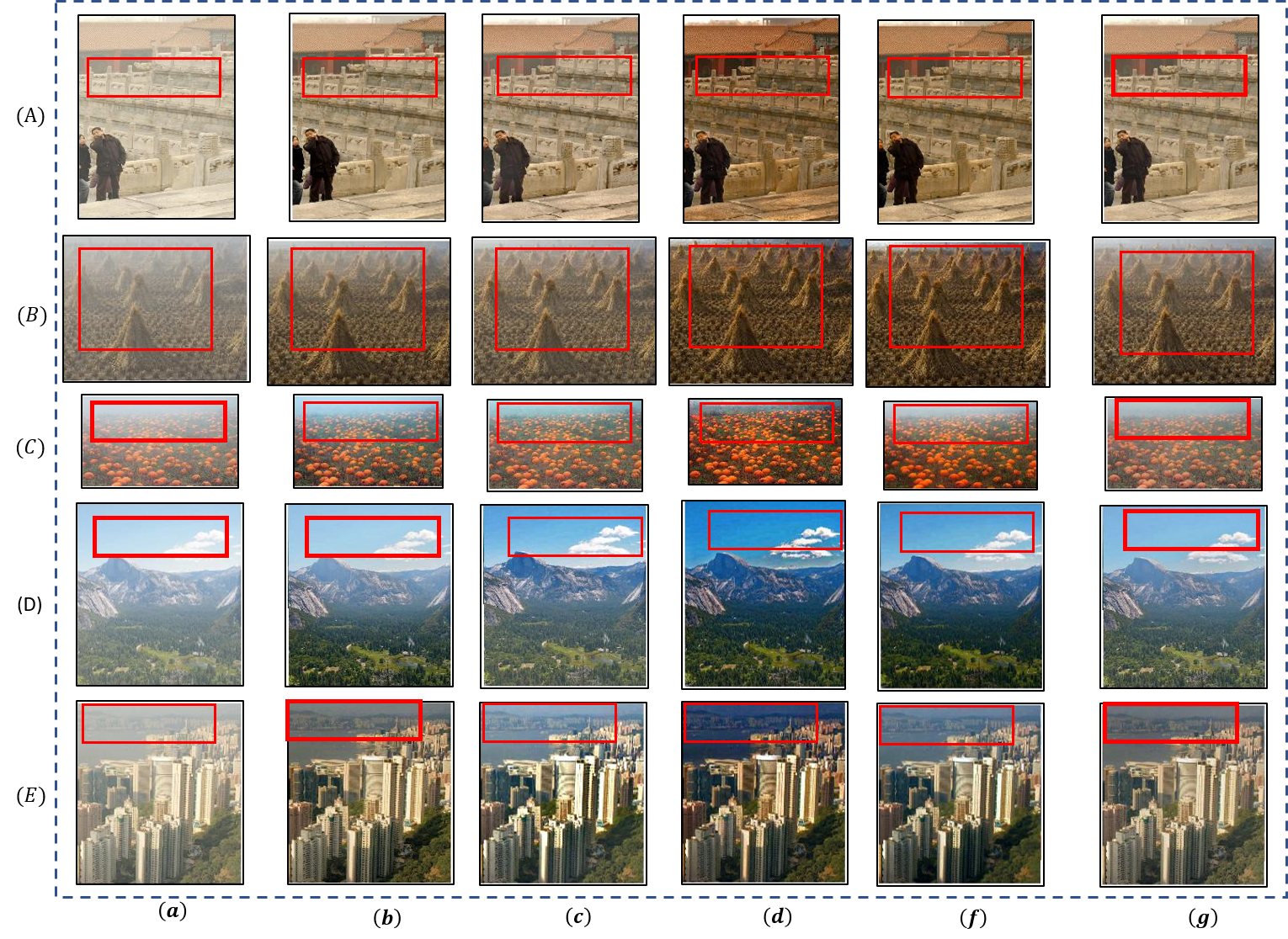}
\end{center}
   \caption{Visual comparison for outdoor real image haze removal. (a) Real world hazy images (b) DehazNet \cite{cai2016dehazenet} (c) Ren \etal \cite{ren2016single} (d) He \etal \cite{he2011single} (e) Zhu \etal \cite{zhu2015fast} (f) Proposed method.}
\label{fig:6}
\end{figure*}
our eyes. Whereas, the proposed method gives better result for all cases (Figure 3 (a-d)) without color distortion and with more haze removal. The reason behind this is \cite{cai2016dehazenet, ren2016single} fail to estimate the scene transmission map accurately in the poor illumination region. Whereas, the proposed C\textsuperscript{2}MSNet is able to estimate the scene transmission map which is very close to the ground truth (\textit{see in Figure 4}).

\subsection{Exp. 2: Outdoor synthetic hazy images}
Performance of the proposed network on outdoor hazy images has been analysed using ImageNet \cite{deng2009imagenet} database. ImageNet \cite{deng2009imagenet} consists verity of classes, among which we considered only two: buildings and beach. Because, these two classes consists considerable variation in natural scenes. Using 3,000 natural scene images from these two classes, we have generated outdoor synthetic hazy image (OHI) database. In this section, we have evaluated the performance of proposed network on generated OHI database. 
\subsubsection{Quantitative analysis}
\par For quantitative analysis, one should have certain basis (ground truth) for the corresponding input. In order to analyse the proposed network on OHI database, transmission maps and synthetic hazy images have been generated as discussed in section 4.3, with the help of Eq. 1. We have used SSIM, MSE and PSNR to compare the proposed  C\textsuperscript{2}MSNet with existing state-of-the-art methods/approaches \cite{he2011single, zhu2015fast, cai2016dehazenet, ren2016single}. Table 3 shows the average SSIM, MSE and PSNR on 3,000 images randomly selected from OHI database. Table 3 witnessed to noticeable increment in the performance evaluation parameters for single image haze removal using proposed method. In natural scenes, constant color shade in sky region causes the color distortion in recovered scene. However, increment in the result witnessed to the usefulness/robustness of cardinal color fusion network to avoid the color distortion.\\  

\begin{table}[t]
\begin{center}
\caption{Quantitative analysis of single image haze removal on OHI dataset.}
\vspace{2mm}
\begin{tabular}{|l|c c c|}
\hline
Method & SSIM & MSE & PSNR \\
\hline\hline
He \etal \cite{he2011single} & 0.8168 & 0.016 & 19.3889 \\
Zhu \etal \cite{zhu2015fast} & 0.7641 & 0.046 & 14.1000 \\
Cai \etal \cite{cai2016dehazenet} & 0.7644 & 0.0484 & 13.8582\\
Ren \etal \cite{ren2016single} & 0.7196 & 0.0676 & 12.4239 \\
\textbf{Proposed method}  & \textbf{0.8910} & \textbf{0.0069} & \textbf{23.7937}\\
\hline
\end{tabular}
\end{center}
\end{table}
\vspace{-5mm}
\subsubsection{Qualitative analysis}
\par Almost all existing methods in recent publications, gives acceptable result on outdoor hazy images (frequently used). It is difficult to sort them in descending order with respect to removal of haze and visual quality. So, to compare with existing dehazing methods, we have employed new set of synthetic hazy images from OHI database. These images comprised of constant shade of cardinal colors. So that, color distortion using existing methods can be easily observed. Figure 5 shows the visual comparison for outdoor image haze removal. As shown in Figure 5 (C,c-f), we can observe that the color distortion in the recovered scene by existing state-of-the-art methods \cite{cai2016dehazenet, ren2016single, he2011single, zhu2015fast}. Whereas, Figure 5 (A-F, g) (proposed method) shows the restored original scene without any color distortion.   
\subsection{Exp. 3: Real world hazy images}
\subsubsection{Qualitative analysis}
In this section, we have carried out only qualitative analysis. Because, it is very difficult to capture the photo of the same scene in hazy and haze free environment. So, in absence of haze free scene (ground truth) we could not conduct the quantitative analysis. For visual (qualitative) analysis, we have considered set of five frequently used (in recent publications) real hazy images. Figure 6 shows the visual comparison for haze removal from real hazy images. From Figure 6, it is observed that the DChP and CAP have more color distortion than the CNN approach. Reason behind this could be the failure of hazy model assumption. The proposed network overcomes this drawback by learning the haze spread.

\section{Conclusion}
In this paper, we proposed a novel C\textsuperscript{2}MSNet for single image haze removal which overcomes the color distortion problem. Proposed network fuses the cardinal color channel and later estimates the transmission map using proposed multi-channel multi-scale CNN. To prove the robustness of proposed network, three databases namely: D-HAZY, OHI and set of real hazy images have been used. Quantitative analysis has been carried out using SSIM, MSE and PSNR. The proposed method also compared with the existing state-of-the-art methods for single image haze removal using visual analysis. Extensive analysis with the help of three experiments shows that the proposed network outperforms the other existing methods for single image haze removal.

\section*{Acknowledgement}
We would like to thank Mr. Sachin Chaudhary and Mr. Prashant Patil, research scholar from Computer Vision and Pattern Recognition (CVPR) Lab, IIT Ropar and Mr. Gajanan Galshetwar for their valuable suggestions in this work. We also thank to anonymous reviewers for their insightful comments and helpful suggestions to improve the quality of this paper. 


\end{document}